# Towards Autonomous Robotic Kidney Ultrasound: Spatial-Efficient Volumetric Imaging via Template Guided Optimal Pivoting

Xihan Ma, Haichong K. Zhang, *Member, IEEE*

*Abstract*—Medical ultrasound (US) imaging is a frontline tool for the diagnosis of kidney related diseases. Current clinical practice is shifting from two-dimensional (2D) imaging towards three-dimensional (3D) volumetric analysis. However, traditional freehand imaging procedure suffers from inconsistent, operator-dependent outcomes, lack of 3D localization information, and risks of work-related musculoskeletal disorders. While robotic ultrasound (RUS) systems offer the potential for standardized, operator-independent 3D kidney data acquisition, the existing scanning methods lack the ability to determine the optimal imaging window for efficient imaging. As a result, the scan is often blindly performed with excessive probe footprint, which frequently leads to acoustic shadowing and incomplete organ coverage. Consequently, there is a critical need for a "spatially efficient" imaging technique that can maximize the kidney coverage through minimum probe footprint. Here, we propose an autonomous workflow to achieve efficient kidney imaging via template-guided optimal pivoting. The system first performs an explorative imaging to generate partial observations of the kidney. This data is then registered to a kidney template to estimate the organ's pose. With the kidney localized, the robot executes a fixed-point pivoting sweep where the imaging plane is aligned with the kidney's long axis to minimize the probe translation. The proposed method was validated in simulation and in-vivo. Simulation results indicate that a 60% exploration ratio provides optimal balance between kidney localization accuracy and scanning efficiency. In-vivo evaluation on two male subjects demonstrates a kidney localization accuracy up to 7.36 mm and 13.84 degrees. Moreover, the optimal pivoting approach shortened the probe footprint by around 75 mm when compared with the baselines. These results valid our approach of leveraging anatomical templates to align the probe optimally for volumetric sweep.

## I. INTRODUCTION

MEDICAL ultrasound (US) imaging remains a cornerstone of modern diagnostic medicine for its non-invasiveness, real-time capability, portability, and non-ionizing radiation [1]. With over 850 million people worldwide affected by kidney diseases [2], renal US serves as the primary frontline tool for detecting many abnormalities such as hydronephrosis, cysts, and complex masses [3]. While renal US traditionally relies on two-dimensional (2D) cross-sections, its clinical usage is increasingly shifting toward three-dimensional (3D) volumetric analysis in recent years, emphasizing whole-organ coverage. For instance, 3D renal US has emerged as the critical tool for the diagnosis and longitudinal monitoring of Autosomal Dominant Polycystic Kidney Disease (ADPKD) [4], allowing clinicians to track kidney size change and cyst progression. In post-transplant renal surveillance, a complete 3D US sweep is becoming necessary for identifying early signs of graft rejection [5] or localized fluid collections [6].

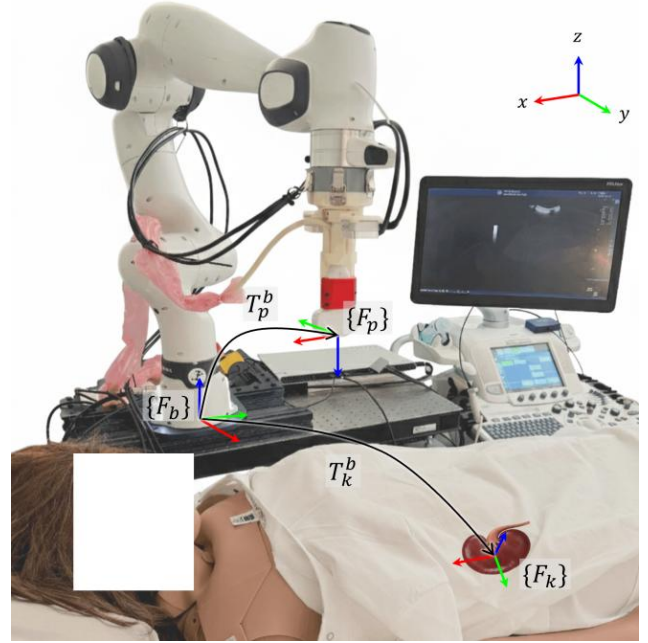

**Fig. 1.** RUS system for autonomous kidney imaging. $\{F_b\}$ denotes the robot base coordinate frame. $\{F_p\}$ denotes the robot end-effector coordinate frame, located at the tip of the US probe. $\{F_k\}$ denotes the canonical frame of the kidney. $T_p^b$ is the rigid transformation from the robot base frame to the end-effector / US image frame. $T_k^b$ represents the rigid transformation from the robot base frame to the canonical kidney frame.

The significance of 3D whole organ imaging lies in (i) the elimination of blind spots in a typical 2D exam; and (ii) more accurate volume measurement which are often used as a diagnostic biomarker [7]. Failure to achieve total volumetric coverage can result in the omission of small, peripherally located lesions or the mis-estimation of pathological changes, compromising the diagnostic reliability.

Despite its importance, current freehand US procedure has significant disadvantages. First, the scanning is highly operator dependent, leading to inconsistent imaging outcomes. Studies have shown ±18% Limits of Agreement (LoA) in renal volume estimation, largely due to variations in manual probe positioning [8]. Moreover, 3D US probes often have limited elevational field-of-view (FOV), leading to difficulties in the complete coverage of the anatomy. Last but not least, the repetitive scanning process can be physically demanding, with up to 90% of sonographers reported work-related musculoskeletal disorders due to non-ergonomic postures and high contact force required [9].

To address these drawbacks, Robotic Ultrasound (RUS) systems that can automate the scanning process have been introduced [10]. By utilizing a robot manipulator to hold the US probe, these systems provide high-precision, wide-area, consistent imaging while eliminating human fatigue. However, achieving autonomous renal imaging remains a technical hurdle. The core challenge lies in the development of (i) an intelligent organ localization method to determine the optimal imaging plane and (ii) a scan strategy to ensure efficient whole-kidney coverage. Without robust methods for organ localization and scan path optimization, the robot may sweep through empty spaces or miss critical portions due to acoustic shadowing, resulting in incomplete volume reconstruction that fail to meet stringent clinical decision-making standards.

### A. Related Work

RUS systems have been used to automate diagnostic imaging for a wide variety of anatomies that require 3D volumetric imaging, including vessels, thyroid, spine, liver, and fetus [10], etc. Autonomy is often realized through the integration of three components: (i) A body surface adaptation method that maintains contact between the probe and the skin for stable acoustic coupling; (ii) A localization algorithm that enables the identification of the desired imaging plane; (iii) A path-planning strategy to build 3D volume.

#### 1) Body Surface Adaptation

Body surface adaptation is a relatively well-studied problem. Both software and hardware-based solutions have been presented to enable contact force and probe orientation control. For example, Jiang et. al. [11] and Huang et. al. [12] demonstrated the use of robot-integrated force/torque sensor to estimate the orthogonal direction of the probe with respect to the body while maintaining a constant contact force. Tsumura et. al. [13] and Lindenroth et. al. [14] proposed to use passive mechanical structures to simultaneously regulate contact force and ensure probe-skin attachment. Ma et. al. showed a rangefinder-integrated end-effector design for real-time probe normal positioning [15], [16].

#### 2) Imaging Plane Localization

Numerous efforts have been made to achieve automatic localization of the desired imaging plane. Among them, Image-based Visual Servoing (IBVS) and Reinforcement Learning (RL) are two common approaches. IBVS aligns the current imaging plane with the desired one by moving tracked features in the image space. Abolmaesumi et. al. [17] and Nadeau et. al. [18] demonstrated IBVS based imaging plane localization using simple image intensity features for thyroid and abdomen imaging, respectively. Mebarki et. al. proposed to use 2D moments of the segmented anatomy as the visual feature which enables simultaneous in-plane and out-of-plane probe control when imaging ellipsoid-shaped anatomies such as kidney [19]. Ma et. al. demonstrated more robust in-plane IBVS using anatomical landmark features for in-vivo lung imaging [20]. Although these methods are straightforward, explainable, and ensure robot control stability, they involve explicit parameterization of visual features which limit the generalizability over different anatomies. Moreover, there is no guarantee that the visual features will remain visible throughout the scan, further hindering its real-world clinical applicability. On the other hand, RL-based approaches do not require cumbersome handcrafted parameterization. It learns a policy by trial-and-erring in simulation, which predicts the next probe movement towards the desired imaging plane. Li et. al. first demonstrated automatic imaging plane localization for lumber spine in a simulated environment [21]. Ning et. al. successfully deployed RL-based probe navigation on phantom setups [22]. Bi et. al. presented a novel anatomy-aware reward function for training RL agents to maximize US wave penetration [23]. They later proposed a RL framework with reward predicting network and was validated on ex-vivo samples [24]. However, it is often difficult to adopt the trained policy to imaging human subjects due to excessively complicated real-world state-space, known as the "real2sim" gap. Furthermore, the trained RL agent operates as a "black box", raising safety concerns on clinical usage. Therefore, it is necessary to find a middle ground approach that is not overly dependent on task-specific parameterization while being deployable and interpretable.

#### 3) 3D Volumetric Scanning

Most previous works adopted straight-line or raster scan strategies for full-coverage volumetric imaging. For example, Jiang et. al. adopted a pipeline that transfers the preplanned straight-line path to the subject by registering intra-operative patient pointcloud to the pre-operative atlas [25]. Many groups have implemented raster scan patterns for the full coverage of neck [26], chest [27], and back [28], etc. Although such scan patterns can achieve full coverage, the trajectory parameters need manual tuning for inter- and intra-patient usage. More importantly, they often result in large trajectory footprints when sliding across the skin. This can be problematic for kidney imaging because the acoustic window is narrower. For such applications, a more spatially efficient strategy that minimizes the probe translational movement should be encouraged.

### B. Contributions

Existing works largely treated imaging plane localization and volumetric scanning as two separate problems. However, 3D

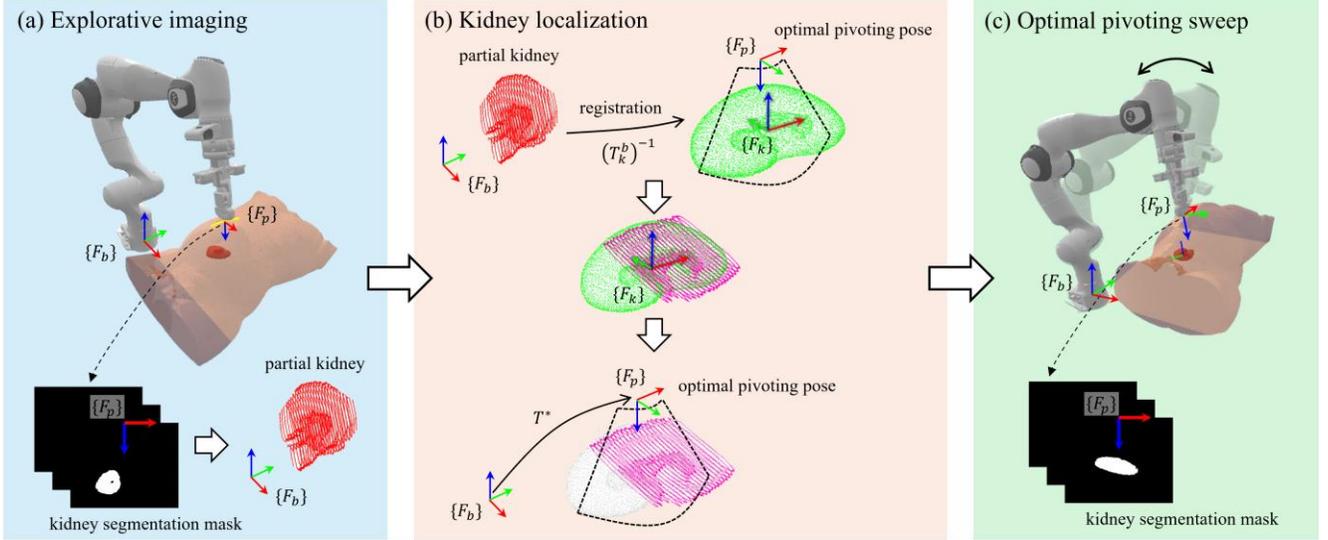

**Fig. 2.** Proposed template-guided, exploration–localization framework for robotic kidney US imaging. The framework involves: **(a)** the explorative imaging phase where the robot scans along a straight-line trajectory (yellow line) to collect US images of a subset of the kidney geometry. A partial kidney point cloud can be built using the spatially localized, kidney segmented US data; **(b)** the kidney localization phase where the partial kidney point cloud is registered to a kidney template model. The registration outcome is used to determine an optimal imaging pose, at which the imaging plane slices the longitudinal principal axis of the kidney; **(c)** the optimal pivoting sweep phase where the robot controls the probe to the optimal pivoting pose determined in the previous phase and performs a sweep with the probe tip position fixed, covering the entire volume of the kidney.

kidney imaging requires a unified pipeline that localizes an imaging plane for optimal volumetric scanning by considering spatial orientation of the organ. To bridge this gap, we propose a novel framework that integrates exploration, localization and optimized scanning. Unlike the "black-box" RL or the anatomy-agnostic raster scan, our method employs a fully interpretable, template guided imaging plane localization approach to facilitate a spatially efficient pivoting scan strategy. Our contributions can be summarized as follows:

- **A Template-Guided Exploration–Localization Workflow**: We propose a multi-stage workflow that derives global kidney pose from partial observations. By registering initial scans to a canonical template, the system estimates patient-specific kidney long-axis orientation, enabling the robot to automatically align the US imaging plane for maximum spatial efficiency during volumetric imaging.

- **A Spatial-Efficient Volumetric Imaging Strategy**: We introduce a fixed-point, pivoting scan strategy that performs a fan-like sweep centered on the kidney's long axis estimated from the previous step. By constraining the probe tip motion, this strategy achieves full kidney coverage with minimum probe translation. This reduces the chance of acoustic shadowing and maximizes the organ volume captured per angular increment.

- **Experimental Validation and Performance Benchmarking**: We demonstrate through a simulation study that an exploration rate of 60% provides robust kidney localization accuracy. Furthermore, we validate the superiority of the proposed optimal pivoting method over translational imaging and non-optimal pivoting sweep. Results from both simulation and human subject trials confirm significant improvements in volumetric reconstruction efficiency and a reduced acoustic window footprint, while maintaining high reconstruction fidelity.

While this work focuses on the renal anatomy as a representative use case, the proposed template-guided framework is inherently generalizable. The underlying principles of exploration–localization and optimal pivoting sweep are applicable to other small organs or anatomical targets where diagnostic volume must be maximized through a constrained acoustic.

The rest of this paper is organized as follows: section II clarifies the problem statement and delineates the technical approach; section III describes the simulation and human subject experiment setup, as well as the benchmarking metrics; section IV presents the experiment results; section V draws conclusions, discusses limitations and potential future directions.

## II. MATERIALS AND METHODS

This section first defines the problem of autonomous robotic kidney imaging. Next, the concept of template guided, exploration – localization workflow is introduced. Then, each building block of this workflow is explained in detail.

### A. Problem Statement

Here, we provide a formal definition of the autonomous robotic kidney US imaging task. The general objective is to achieve maximum volumetric coverage of the renal anatomy with minimum translational displacement of the US probe. This strategy is determined by the physical constraints of transabdominal imaging. Clinically, the acoustic window (the narrow gap between ribs or bowel gas that allows US waves to reach kidney) is fixed. Excessive probe translation may shift the transducer away from this window, leading to shadowing or loss of US signal. Conversely, maximum volumetric coverage ensures that the entire organ can be evaluated, which is critical for identifying peripheral pathologies that are frequently missed in sparse 2D exams. By minimizing probe translational displacement while maximizing internal visualization, the robot mimics the "sweeping" technique of human sonographers, who maintain a single point of contact to avoid obstructive anatomy.

A kidney is geometrically characterized by its longitudinal principal axis. Intuitively, the US imaging plane must be aligned with this principal long-axis plane, $\Pi_L$, to minimize the angular sweep that encompasses the organ's volume. Therefore, we define an optimal probe pose, $T^* = (R^*, t^*) \in SE(3)$, where $R^* \in SO(3)$, $t^* \in \mathbb{R}^3$, such that: (i) the US imaging plane is coplanar with $\Pi_L$; (ii) when positioning the US probe tip at the fixed point on the skin, $t^*$, an unobstructed, kidney-centered image is acquired. Under these conditions, the most efficient trajectory is a pivoting motion at $t^*$, where the robot rotates the probe about its $x$-axis. This strategy minimizes both the probe translational motion and the likelihood of internal acoustic occlusion.

The central challenge in achieving this strategy is that $\Pi_L$ is located differently for each patient and unknown to the robot. Therefore, we treat the imaging task as an anatomical localization problem: Let $\{F_b\}$ be the robot base frame, $\{F_k\}$ be a canonical coordinate frame of the kidney (axes are aligned with the organ's principal dimensions), the task is to estimate a rigid transformation, $T_b^k \in SE(3)$, from $\{F_k\}$ to $\{F_b\}$. The optimal probe pose, $T^*$, can be easily computed once $T_b^k$ is accurately measured.

To focus better on addressing the above challenge, we operate under the following assumptions: (i) The robot is assumed to have initially landed on the abdomen where kidney is visible. This is technically achievable using external vision systems to identify the approximate kidney location and has been validated by existing literature for similar robotic US imaging applications [29], [30], [31]; (ii) We assume a standard clinical "breath-hold" protocol. The kidney is treated as a rigid body during the brief duration of the scanning phase, neglecting respiratory-induced non-rigid deformations; (iii) Clinically, the right kidney is easier to image than the left one because of larger image window provided by the liver. To demonstrate the feasibility of the proposed framework, we discuss only right kidney imaging for the remaining of the paper. Extension to the left kidney and other small organs are discussed in section V.

### B. Technical Approach Overview

We developed an RUS system for the kidney US imaging task. As depicted in Fig. 1, a 7 Degree-of-Freedom (DoF) robotic manipulator (FR3, Franka Emika, Germany) is positioned at the patient's bedside. A customized end-effector holding a curvilinear US probe (C1-6, GE Healthcare, USA) is mounted to the robot through a clamping mechanism. The rigid body transformation from the last link of the robot to the probe tip is calibrated using the end-effector's CAD model. The robot control pipeline is implemented based on the PandaPy framework [32]. This pipeline controls the probe pose through joint velocity commands, allowing the probe to slide over the patient skin surface smoothly and safely (section II-C).

The US probe is connected to a medical grade cart-based US machine (Logiq E9, GE Healthcare, USA). The imaging parameters were configured to the default abdominal setting. A video capture card (Video2USB, ClearClick, USA) is used to grab US frames at 20 frame-per-second (fps). A neural network is used to segment the kidney contour from real-time US images at 20 fps, then streamed through Robot Operating System (ROS) network for image-based feedback control.

A template-guided, exploration–localization framework is proposed to address the challenge of unknown principal long-axis plane of the organ. This framework is illustrated in Fig. 2. To provide a reference for localization, we developed a canonical kidney model, derived from a public CT dataset (section II-D). By applying Principal Component Analysis (PCA) to a population of segmented kidneys, we extract the mean morphology and define a canonical kidney frame $\{F_k\}$. This template serves as the global map for the subsequent registration process. Since the transformation $T_b^k$ is initially unknown, the robot executes an exploration trajectory (section II-E). During this phase, 2D US images with kidney segmented and the corresponding probe poses are collected in real-time. These kidney cross-sections are back-projected into the robot base frame $\{F_b\}$ to form a partial kidney point cloud. Localization is achieved by registering the partial point cloud to the global template. We treat this as a rigid registration problem and use constrained Iterative Closest Point (ICP) algorithm to find the optimal $T_b^k$ that minimizes the spatial discrepancy between the partial observation of the kidney and the canonical model (section II-F). Once $T_b^k$ is estimated, the optimal imaging pose $T^*$ is calculated by projecting the longitudinal axis and the centroid of the template kidney model onto the robot's task space (section II-G). Finally, the robot maneuvers the probe to reach $T^*$ by sliding across the skin surface. A volumetric imaging is executed where the robot maintains the probe tip at the projected centroid and rotates the probe to sweep over the kidney. This motion generates a dense sequence of cross-sections of the entire kidney that can be used for later diagnosis.

### C. US Probe Pose Control

Here, we describe the control architecture which allows us to define the US probe pose while simultaneously ensuring acoustic coupling and avoiding kinematic singularities. We formulate the control of the 7 DoF robotic manipulator using a velocity-based scheme for motion smoothness. Hence, the goal is to solve for the joint velocities $\dot{q} \in \mathbb{R}^7$ to achieve the desired probe maneuvers.

First, we consider the rotational motion of the probe. To facilitate tilting and rotating the US probe, the rotational control is decoupled into two alignment tasks:

- Aligning the $z$-axis of $\{F_p\}$ (the approach vector) to the desired direction, i.e., probe tilting.

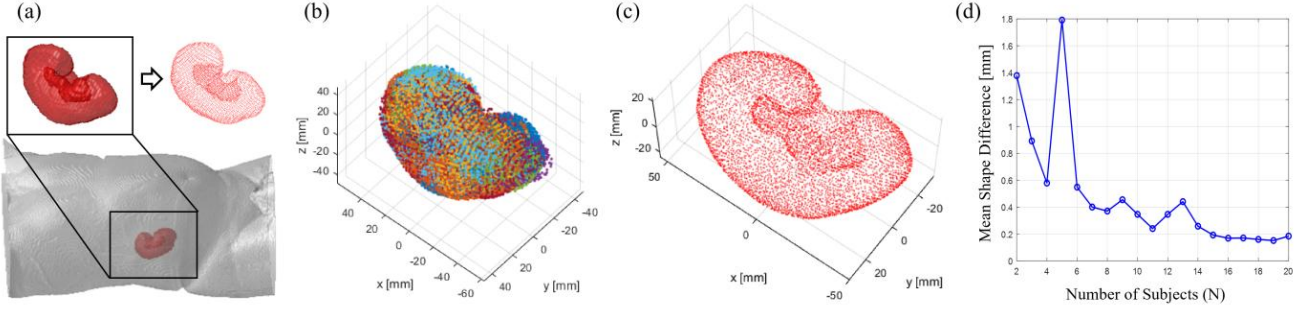

**Fig. 3.** Kidney template model computation. **(a)** The extraction of the kidney mesh from the abdominal CT data and the conversion to point cloud representation for one subject. **(b)** Registered kidneys from all subjects. Each color in the point clouds represent one subject. **(c)** The kidney template model built by averaging corresponded points of all the kidneys. The point cloud is in the canonical kidney coordinate frame where the $x$-axis aligns with the longitudinal principal axis, the $y$-axis aligns with the lateral principal axis. **(d)** The per-point Euclidean difference of the kidney template model when built using different number of subjects.

- Aligning $x$-axis of $\{F_p\}$ (the orientation vector) to the desired direction, i.e., probe rotating.

The error of the approach vector, $e_a$, can be computed as:

$$e_a = a_{curr} \times a_{des} \qquad (1)$$

where $a_{curr} \in \mathbb{R}^3$ is the current approach vector expressed in $\{F_b\}$, extracted from the robot forward kinematics (i.e., $T_p^b$, as illustrated in Fig. 1); $a_{des} \in \mathbb{R}^3$ is the desired approach vector, expressed in $\{F_b\}$. Then, the angular velocity required to align $a_{curr}$ to $a_{des}$ can be computed by:

$$\omega_a = K_{p1}e_a + K_{p11}(a_{curr} \times \dot{a}_{des}) \qquad (2)$$

where $K_{p1}$ and $K_{p11}$ are empirically tuned control gains. Similarly, the orientation vector error, $e_o$, is computed by:

$$e_o = o_{curr} \times o_{des} \qquad (3)$$

where $o_{curr} \in \mathbb{R}^3$ and $o_{des} \in \mathbb{R}^3$ are the current and the desired orientation vector, both expressed in $\{F_b\}$. The angular velocity that aligns $o_{curr}$ to $o_{des}$ without interfering with the previous alignment task can be computed as:

$$\omega_o = K_{p2}(e_o \cdot a_{curr}) \qquad (4)$$

The final desired probe angular velocity is given by:

$$\omega_{total} = \omega_a + \omega_o \qquad (5)$$

Second, we consider the translational motion of the probe. Stable acoustic coupling requires precise regulation of the contact force $F_z$ along the probe's approach vector (z-axis). Hence, we implemented a PD control law to regulate $F_z$:

$$v_z = \left[ K_{p3}(\hat{F}_z - F_z) + K_{d1}\frac{d}{dt}(\hat{F}_z - F_z) \right] a_{curr} \qquad (6)$$

where $v_z$ is the probe linear velocity along its z-axis, expressed in $\{F_b\}$; $K_{p3}$ and $K_{d1}$ are control gains; $\hat{F}_z$ is the desired contact force; $F_z$ is the current contact force expressed in $\{F_p\}$, estimated from joint torque sensor readings using manufacture provided API. The probe linear velocities along its $x$-, and $y$-axis are computed by:

$$\begin{cases} v_x = K_{p4}(\hat{P}_x - P_x) \\ v_y = K_{p5}(\hat{P}_y - P_y) \end{cases} \qquad (7)$$

where $K_{p4}$ and $K_{p5}$ are control gains; $\hat{P}_x$ and $\hat{P}_y$ are the desired probe tip location in $\{F_b\}$, specified by the user. The final desired probe linear velocity is given by:

$$v_{total} = \begin{bmatrix} v_x & v_y & v_z \end{bmatrix}^T \qquad (8)$$

Eventually, the resulting task-space twist expressed in the robot base frame is:

$$\xi_b = [\omega_{total} \; v_{total}]^T \qquad (9)$$

We then solve for the joint velocities $\dot{q}$:

$$\dot{q} = J^\dagger \xi_b + (I - J^\dagger J)\dot{q}_{null} \qquad (10)$$

where $J \in \mathbb{R}^{6 \times 7}$ is the space Jacobian of the robot; $J^\dagger$ is the Moore-Penrose pseudoinverse of the Jacobian; $(I - J^\dagger J)$ is the null-space projector. To avoid the arm from approaching to singular configurations, we utilize the null-space joint velocity $\dot{q}_{null} \in \mathbb{R}^7$ to maximize the arm's manipulability index: $m(q) = \sqrt{\det(JJ^T)}$. The null-space term is calculated using the gradient of the manipulability:

$$\dot{q}_{null} = K_{p6} \nabla_q m(q) \qquad (11)$$

where $q \in \mathbb{R}^7$ is the current joint configuration; $K_{p6}$ is the control gain; the gradient term is computed numerically using a small perturbation value $\epsilon$ for each joint $q_i$:

$$\frac{\partial m}{\partial q_i} \approx \frac{m(q_i + \epsilon) - m(q_i)}{\epsilon} \qquad (12)$$

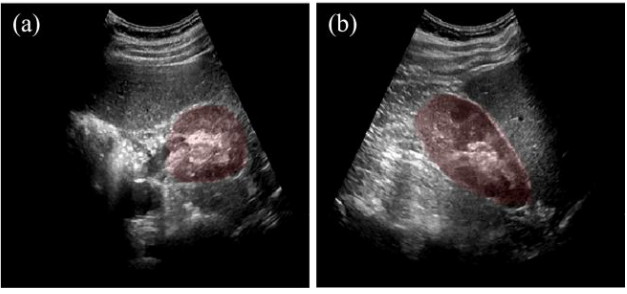

**Fig. 4.** Kidney segmentation examples during **(a)** explorative imaging, and **(b)** pivoting sweep. The red overlay is the segmentation mask.

This task space to joint space mapping ensures that the imaging task is prioritized while the redundant DoF is utilized to keep the robot in a dexterous configuration.

In summary, this control architecture tracks user-intended probe orientation ($a_{des}$, $o_{des}$) and position ($\hat{P}_x$, $\hat{P}_y$) while maintaining constant pressure between the probe and the body. In the following sections, we will use this architecture as a foundation and focus on the calculation of the intended probe pose.

### D. Kidney Template Model

This section details the construction of a statistical kidney template model to serve as the canonical reference for the later localization task. The generation process involves extracting anatomical manifolds from a multi-subject dataset, establishing point-wise correspondences, and defining a robust, constrained coordinate frame.

To capture the inter-patient morphological variation of the renal system, we utilize a public abdominal CT dataset with kidney binary mask labels [33]. For each subject $i \in \{1, \dots, N\}$, the kidney mask $O_i$ is processed using the Marching Cubes algorithm to extract raw surface manifold. We sample the vertices of the resulting manifold to obtain a raw point cloud $P_i \in \mathbb{R}^3$ in the CT coordinate system (see Fig. 3a). We then define a standardized kidney coordinate system $\{F_k\}$ by applying PCA on the raw point cloud. Specifically, eigen decomposition is performed on the covariance matrix of $P_i$:

$$C = \frac{1}{M-1} \sum_{j=1}^{M} (p_j - c_i)(p_j - c_i)^{\mathrm{T}} = V D V^{\mathrm{T}} \qquad (13)$$

where $C$ is the covariance matrix; $M$ is the number of vertices; $c_i$ is the centroid of $P_i$ ; $V = [v_1, v_2, v_3]$ contains the eigenvectors (principal axes of kidney); $D$ is the diagonal matrix of eigenvalues $\lambda_1 \geq \lambda_2 \geq \lambda_3$ . However, we cannot directly use $V$ as the basis for the canonical frame $\{F_k\}$ because of the inherent sign ambiguity of PCA. Thus, constraints are needed to ensure consistent axes orientation across different kidneys. Denote the final canonical basis as $\{x_k, y_k, z_k\}$, the following constraints are enforced sequentially:

- The primary axis must point from inferior to superior of the body (represented as a vector, $x_{ref}$ ), i.e., $x_k = \mathrm{sgn}(v_1 \cdot x_{ref}) \, v_1$.

- The secondary axis must point from medial to lateral of the body (represented as a vector, $y_{ref}$ ), i.e., $y_k = \mathrm{sgn}(v_2 \cdot y_{ref}) \, v_2$.
- The third axis obeys the right-hand rule, i.e., $z_k = x_k \times y_k$, followed by orthonormalization of $y_k$.

The resulting homogenous transformation matrix that transforms points $P_i$ from the CT data coordinate frame to the kidney canonical frame is:

$$T_{CT}^k = \begin{bmatrix} R_i & c_i \\ 0_{1 \times 3} & 1 \end{bmatrix} \qquad (14)$$

where $R_i = [x_k, y_k, z_k]$ . Applying this transformation to all subjects yields a centered, principal axes aligned kidney population which can be used to further compute a statistical mean kidney shape.

Next, we map the point clouds of all the kidneys to a common space using a multi-stage pipeline. First, each point cloud is resampled using Farthest Point Sampling algorithm to ensure uniform sampling. Then, a feature-based registration pipeline introduced in [34] is applied to initially align all kidney point clouds (see Fig. 3b). Lastly, we utilize $k$-Nearest Neighbor (kNN) search in the aligned space to re-index the kidney vertices, ensuring that for every subject $i$, vertex $j$ represents the same anatomical landmark. The final statistical kidney template (see Fig. 3c) is defined by the mean vertex positions across the spatially aligned kidney point clouds:

$$P_{template} = \frac{1}{N} \sum_{i=1}^{N} P_i^{aligned} \qquad (15)$$

We used $N = 20$ subjects to compute the kidney template. To ensure this sample size is sufficient to represent inter-patient kidney morphology, we performed a convergence analysis by calculating the mean difference between $P_{template}$ when using $1, 2, \dots, 20$ subjects. The result shows that $P_{template}$ difference decreases exponentially as the number of subjects increases and reaches a steady state when $N = 20$ (see Fig. 3d). This confirms that our sample size is sufficiently large.

### E. Initial Explorative Imaging

Explorative imaging is needed to acquire partial information about the patient's kidney for the later localization step. To start the exploration process, the probe is first manually positioned on the patient's abdomen where the kidney is visible in the 2D US stream. Although literature has confirmed that automatic probe placement is possible, we performed this process manually simplify the system and better focus on the kidney localization problem.

The exploration is performed by sliding the probe from the superior to the inferior of the patient, where the US images capture the transverse slices of the kidney. This scan strategy allows the US probe to slide over a relatively flat body surface where high contrast images can be obtained more easily.

Real-time semantic segmentation is performed using the YOLO11n-seg model [35]. We chose this model for its ease to train and high inference speed. Based on the pretrained version of the model, we fine-tuned it on a composite dataset which contains 168 kidney US images from a public dataset [36] and 154 US images acquired in-house using our US machine. Training labels were generated semi-autonomously using a large language model, Med SAM [37]. The YOLO11n-seg model was trained using the loss function below:

$$L_{\text{loss}} = \lambda_{\text{cls}}L_{\text{cls}} + \lambda_{\text{box}}L_{\text{box}} + \lambda_{\text{msk}}L_{\text{msk}} \qquad (16)$$

where $L_{\text{cls}}$ is the binary cross-entropy loss to correctly identify kidney over background; $L_{\text{box}}$ is the Intersection over Union (IoU) loss to help localize the kidney; $L_{\text{msk}}$ is the Dice loss to ensure the per-pixel segmentation accuracy; $\lambda_{\text{cls}}$, $\lambda_{\text{box}}$, and $\lambda_{\text{msk}}$ are scalar weights. We use 80% of the data for training and 20% for testing. The testing accuracy after training for 50 epochs is 80.8% in terms of IoU. Kidney segmentation examples can be found in Fig. 4.

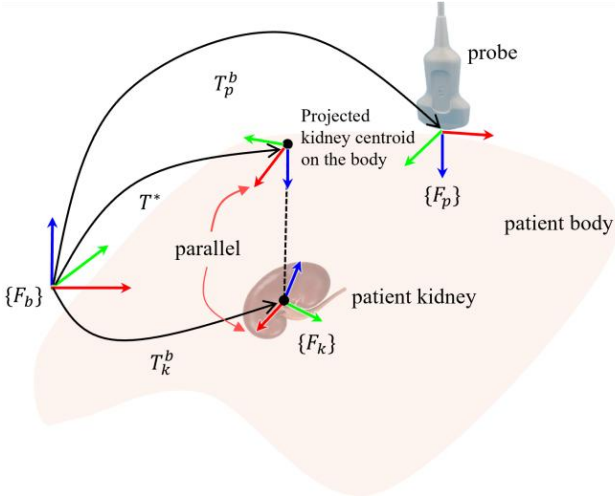

**Fig. 5.** Illustration of the optimal pivoting pose. $T^*$ depicts the transformation from the robot base to the optimal pivoting pose. To capture the longitudinal principal axis of the kidney, the $x$-axis of the optimal pivoting pose is parallel to the $x$-axis of the estimated canonical kidney coordinate frame; the origin of the optimal pivoting pose is a projection of the kidney canonical frame's origin onto the body along the $z$-axis of the robot base frame $\{F_b\}$.

Once the kidney is detected, the robot executes a straight-line exploration trajectory across the body surface. To prevent the kidney from exiting the FOV, we implement a controller that centers the organ in the 2D US image while moving. The controller is designed based on the architecture in section II-C. During exploration, the desired probe orientation ($a_{\text{des}}$, $o_{\text{des}}$) is kept unchanged, whereas the desired probe position at each timestamp $t + \Delta t$ is given by:

$$\begin{cases} \widehat{P}_x(t + \Delta t) = \widehat{P}_x(t) + (\delta_{\text{msk}}\Delta t) \cdot (\hat{x}_p \cdot [1,0,0]^{\text{T}}) \\ \widehat{P}_y(t + \Delta t) = \widehat{P}_y(t) + v_{\text{scan}}\Delta t \end{cases} \qquad (17)$$

where $\Delta t$ is the control interval; $v_{\text{scan}}$ is a constant scan speed, set to 5 mm/s; $\delta_{\text{msk}} = K_{p7}(c_{\text{msk}} - 0.5)$ centers the kidney mask in the image space ($K_{p7}$ is the control gain, $c_{\text{msk}} \in [0,1]$ is the horizontal centroid of the kidney mask); $\hat{x}_p = R_p^b[1,0,0]^{\text{T}}$ is the probe's $x$-axis in the robot base frame ($R_p^b$ is the rotational component of $T_p^b$.). The probe's $z$-axis motion is automatically controlled to maintain the contact force. Meanwhile, the patient is asked to breath-in and hold so that the kidney remains static with respect to the robot.

As the probe slides on the patient's body, the system records temporal sequence of probe tip poses $T_{p,t}^b$ and the corresponding kidney binary masks at 20 fps. The boundary pixels of the kidney mask are back projected to the robot base frame using $T_{p,t}^b$, generating a point cloud $P_{\text{local}} \in \mathbb{R}^3$ that represents the local observation of the kidney.

### F. Patient Kidney Localization

The objective of the localization phase is to estimate the rigid transformation $T_k^b$ that maps the kidney's canonical frame $\{F_k\}$ to the robot base frame $\{F_b\}$. While the canonical frame can be defined using PCA on complete CT volumes (section II-D), it cannot be reliably computed directly from the explorative point cloud $P_{\text{local}}$. This is because the US images acquired from the patients are inherently noisy (so as the segmentation) and only represent a subset of the organ's surface. Using PCA on such data would yield biased principal axes. However, we can estimate the patient kidney's canonical coordinate by registering $P_{\text{local}}$ to the kidney template point cloud, $P_{\text{template}}$. Therefore, we treat the localization as a registration problem where we use the kidney template to regularize the patient's kidney pose.

The challenge of this registration problem lies in the following aspects: (i) $P_{\text{local}}$ only contains partial kidney; (ii) The kidney has smooth surface without unique anatomical landmarks; (iii) The shape of the patient kidney can be significantly different from the template kidney. Therefore, the registration is inherently underdetermined. To overcome this challenge, we use an iterative registration algorithm, ICP, with heuristic initial alignment, customized cost function, and optimization step-size constraints.

We first perform a heuristic initial alignment utilizing known patient posture and the robot's bedside orientation. This step avoids sign-flipping and axis-swapping ambiguities caused by the incomplete explorative kidney shape. Specifically, since the patient always lies in the longitudinal bedside direction (see Fig. 1), we formulate the initial rotation $R_0 \in SO(3)$ by making sure the $x$-axis of the template kidney is parallel to the $y$-axis of the robot base frame. Next, the initial translation $t_0$ is set to $[0,0,0]^{\text{T}}$.

Next, we refine the initial alignment by employing the ICP algorithm using the plane-to-plane cost function. Unlike regular point-to-point cost function, our approach minimizes the distance between local tangent planes of both point clouds, significantly improving registration accuracy on smooth, feature-sparse anatomical surfaces. At each iteration, for a transformed point $p_i$ in $P_{\text{local}}$, we find its nearest neighbor $q_j$ in $P_{\text{template}}$ identified using kNN search, as well as the local surface normal $n_j$. The plane-to-plane loss function minimizes the spatial displacement error projected onto the surface normal of the template:

$$\arg\min \sum_{i=1}^{M} \left( n_j^{\mathrm{T}}(Rp_i + t - q_j) \right)^2 \qquad (18)$$

where $M$ is the number of points in $P_{\mathrm{local}}$; $R$ and $t$ are the rotation matrix and translation vector to be optimized. We solve this optimization using the small-angle approximation: $R \approx I + [\delta\theta]_\times$, where $[\delta\theta]_\times$ is the skew-symmetric matrix of the incremental rotation $\delta\theta = [\delta\theta_x, \delta\theta_y, \delta\theta_z]^{\mathrm{T}}$. This way, the loss function is linearized into the form of $Ax = b$, where the state vector $x = [\delta\theta^{\mathrm{T}}, dt^{\mathrm{T}}]^{\mathrm{T}}$ containing the incremental rotation and translation updates, respectively. A critical feature of our implementation is the restricted rotation updates. To prevent over-rotation during the optimization, we apply the following clamping constraint to the incremental rotation $\delta\theta$:

$$\delta\theta = \max(-\theta_{\max}, \min(\delta\theta, \theta_{\max})) \qquad (19)$$

By restricting the rotation per iteration, the system maintains the global anatomical orientation established by the initial heuristic alignment while allowing fine adjustment of the explorative kidney to match the template. The iteration terminates when the change in the projection error is below $10^{-5}$, yielding the final estimated transformation $T_k^b$.

### G. Optimal Pivoting for Whole Kidney Imaging

The final stage of the framework utilizes the localized kidney pose to execute a volumetric sweep. This section describes (i) the derivation of the optimal pivoting pose $T^*$; (ii) the probe transition from where the explorative imaging ended to the optimal pivoting pose; and (iii) the US-guided sweeping strategy.

First, we determine the optimal pivoting pose $T^*$ that achieves maximum volumetric coverage with minimum probe translational displacement. While the ideal pose would align the US imaging plane exactly with the kidney's longitudinal principal plane $\Pi_L$ (i.e., the x-z plane of $\{F_k\}$), such a pose could result in extreme probe tilting angle that exceeds the physical constraints of the acoustic window or the robot's kinematics capability. Therefore, we relax the requirement of alignment by only enforcing the transducer array (the probe's x-axis) to be parallel with the estimated organ long axis (the kidney's x-axis). Denote the rotation and translation components of $T_k^b$ to be $R_k^b \in \mathrm{SO}(3)$ and $t_k^b \in \mathbb{R}^3$, respectively. Consider the relaxed $T^*$ in the following form:

$$T^* = \begin{bmatrix} r_1^* & r_2^* & r_3^* & p^* \\ 0 & 0 & 0 & 1 \end{bmatrix} \qquad (20)$$

Frist, we construct the rotational component: Set $r_3^* = [0,0,-1]^{\mathrm{T}}$ to make the probe point downwards. Then, align $r_1^*$ to the x-axis of the kidney canonical frame:

$$r_1^* = \mathrm{unit}(R_k^b[1,0,0]^{\mathrm{T}} - (R_k^b[1,0,0]^{\mathrm{T}} \cdot r_3^*)r_3^*) \qquad (21)$$

where $\mathrm{unit}(\cdot)$ represents the vector normalization operation. Lastly,

$$r_2^* = \mathrm{unit}(r_1^* \times r_3^*) \qquad (22)$$

Second, the translational component is set to: $p^* = t_k^b$. This way the US imaging plane is still guaranteed to slice $\Pi_L$ while the probe tilting angle remained minimum for stable contact. A graphical representation of $T^*$ can be found in Fig. 5.

To move the probe to $T^*$, we again, use the control architecture in section II-C. This time, the desired probe orientation is $(a_{\mathrm{des}} = r_3^*, o_{\mathrm{des}} = r_1^*)$. Given $p^* = [p_x^*, p_y^*, p_z^*]^{\mathrm{T}}$, the desired probe position is $(\hat{P}_x = p_x^*, \hat{P}_y = p_y^*)$. Note that with the automatic contact force control, the robot will slide the probe to $(p_x^*, p_y^*)$ over the patient body. Hence, $p_z^*$ does not affect the process of reaching the optimal pivoting pose.

Once $T^*$ is reached, the robot executes a fixed-point pivoting sweep. The desired probe tip position ($\hat{P}_x$, $\hat{P}_y$) and the orientation vector ($o_{\mathrm{des}}$) will stay unchanged during the sweep, whereas the approach vector $a_{\mathrm{des}}$ rotates about $o_{\mathrm{des}}$. At each timestamp $t + \Delta t$, the desired approach vector is updated using the Rodrigues formula:

$$a_{\mathrm{des}}(t + \Delta t) = a_{\mathrm{des}}(t)\cos(\Delta\theta) \\ + (o_{\mathrm{des}} \times a_{\mathrm{des}}(t))\sin(\Delta\theta) \qquad (23)$$

where $\Delta\theta = \pm\dot{\theta}\Delta t$ determines the desired incremental tilt, and the sign determines the sweep direction (clockwise or counter clockwise); $\dot{\theta}$ is a fixed angular rate. Because $a_{\mathrm{des}}$ and $o_{\mathrm{des}}$ are orthogonal, the third term in the Rodrigues formula involving $a_{\mathrm{des}} \cdot o_{\mathrm{des}}$ is omitted. The pivoting sweep utilizes the real-time kidney segmentation to bound the motion: The robot first rotates in clockwise until the kidney disappears from the FOV, then rotates in the reversed direction to capture the other half of the kidney until it is invisible again. With this pivoting motion, the whole kidney volume is captured through a single acoustic window.

### III. EXPERIMENT VALIDATION

This section describes the experiment design to validate the proposed robotic US kidney imaging framework. First, we performed a preliminary simulation study to determine the exploration coverage needed to achieve stable kidney localization. Next, we conducted both simulation and in-vivo experiments to evaluate the system's performance in terms of the kidney localization accuracy and the optimal pivoting efficiency.

### A. Simulation Environment Setup

We developed a simulation environment to benchmark the proposed framework with absolute geometric ground truth. The simulation environment is built on the Pybullet framework. For the sake of simplicity, soft-tissue deformation and acoustic effects (e.g., speckles, artifacts, shadowing, etc.) are not considered. Our RUS system model along with virtual patient models are loaded into the environment (see Fig. 6a).

To generate realistic patients, we extracted the body and kidney boundaries from the CT dataset introduced in section II-D and converted them into triangular meshes. The extraction of the body adopts the same workflow in [38]. To emulate real-time US feedback, we implemented an oblique slicing pipeline. At each control cycle, the probe's 6-DoF pose defines a clipping plane that intersects the patient's CT volume. This allows us to generate a 2D image replicating the FOV of an US image, containing labeled kidney and other surrounding anatomies. A total of number of 15 virtual patients (different from the subjects used for building the kidney template model) are employed for simulation validation.

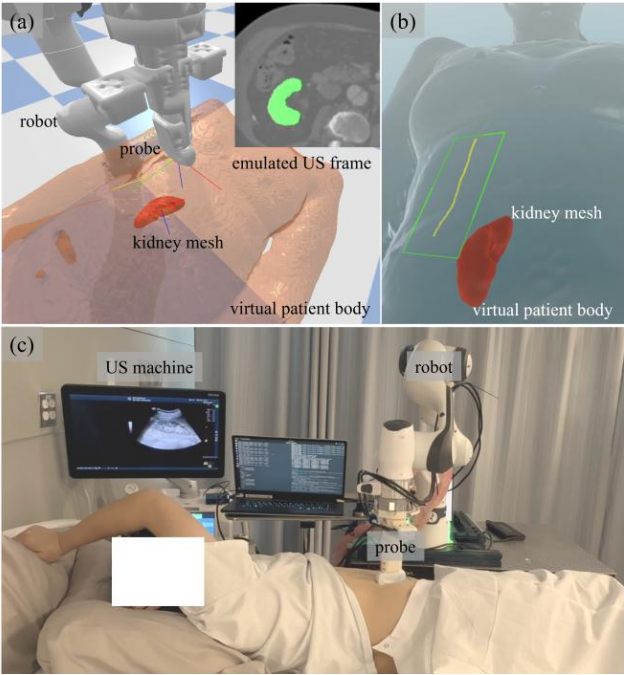

**Fig. 6.** Experiment setup. **(a)** Simulated robotic kidney imaging scene. The robot is performing explorative imaging within the Exploration Region of Interest (E-ROI). The real-time emulated US image with kidney segmentation (green overlay) is shown on the top right corner. **(b)** Generation of the E-ROI (green rectangular region) by projecting the kidney mesh onto the patient body. An example exploration trajectory (yellow line) is generated within the E-ROI. **(c)** In-vivo experiment setup.

### B. In-vivo Experiment Setup

Two healthy male subjects are recruited for the in-vivo study. They exhibit different body shapes (BMIs: 26.2 and 16.9) and allow us to test the system on subjects with anatomical variability. The hardware RUS system is deployed at the subject's bedside (see Fig. 6c). The subjects are positioned to the left lateral recumbent posture to better expose the right kidney. This posture is aligned with the current clinical practice [39]. A breath-hold protocol was conducted to minimize motion artifacts during scanning. Subjects were instructed to breath-in and hold during the explorative imaging step and the optimal pivoting sweep step. The duration of each step is less than 20 seconds. Normal respiration is permitted during the kidney localization phase where the ICP-based registration is performed in the background. During this phase, there is no active probe motion except for maintaining the constant contact force. We empirically set the desired contact force to 9 N, which allows for clear kidney visualization without causing major discomfort to the subjects.

### C. Preliminary Study: Investigate Exploration Coverage vs. Kidney Localization Accuracy in Simulation

Since the kidney localization is an underdetermined problem (as explained in section II-F), a fundamental trade-off in our proposed framework is the exploration coverage vs. the kidney localization uncertainty. While capturing too little of the patient's kidney can lead to poor registration results, extensive exploration (e.g., 100% of the kidney) would reduce the efficiency of the procedure. Thus, the goal of this experiment is to determine the "sweep-spot" exploration coverage that balances localization accuracy and procedural efficiency. This study can only be conducted in simulation because it requires the ground-truth shape of the kidney (to build the canonical frame) and the ground-truth kidney pose $T_{k,\text{gt}}^b$ relative to the robot.

In the simulation environment, we first define an Exploration Region of Interest (E-ROI) by projecting the cuboid bounding box of the kidney mesh vertically onto the patient's abdominal surface (see Fig. 6b). Within the E-ROI, we simulate straight-line trajectories along the $y$-axis of $\{F_b\}$. The robot then explores the kidney along this trajectory in one direction. We define the Exploration Ratio (ER $\in [0,1]$) as the fraction of total kidney length traversed by the probe. For each virtual patient, we increase ER from 0% to 100% at a 10% interval. At ER = 0%, there is no information about the patient's kidney, and the localization is purely based on the initial alignment. At ER = 100%, the entire kidney shape is obtained.

With the partial kidney point cloud obtained with different ER, we estimate the kidney pose with respect to the robot base, $T_{k,\text{est}}^b$, using the registration pipeline in section II-F. The ground truth kidney pose can be retrieved from the simulation. We then evaluate he kidney pose estimation accuracy by computing the difference between $T_{k,\text{est}}^b$ and the ground truth transformation $T_{k,\text{gt}}^b$ in terms of translation and rotation.

- The translation error ($E_{\text{trans}}$): The Euclidean distance in $x$-, and $y$-axis between the estimated and the ground truth kidney centroid.

- The rotation error ($E_{\text{rot,x}}$): The absolute angular difference between the estimated and the ground truth $x$-axis of the kidney canonical coordinate.

The reason to only evaluate the $x$-axis difference is explained by (20) to (22) in section II-G. We use the smallest ER at which both the translation and rotation error converge as the critical ER for all subsequent simulation and in-vivo experiments.

### D. Validate Kidney Localization Accuracy in Simulation

The objective of this experiment is to quantify the kidney localization accuracy under controlled conditions. By utilizing the simulation environment, we can establish an absolute baseline for our framework's ability to recover the kidney's canonical frame from partial observation data.

For each virtual patient, the robot executes the exploration phase using the critical ER determined in section III-C. The process of generating the exploration trajectory is also the same as described in section III-C. To ensure the results are not biased by the probe's initial position, we perform five randomized exploration runs per subject. Each run originates from a different starting point within the E-ROI. The starting points are selected such that there is enough distance left to complete the required path length.

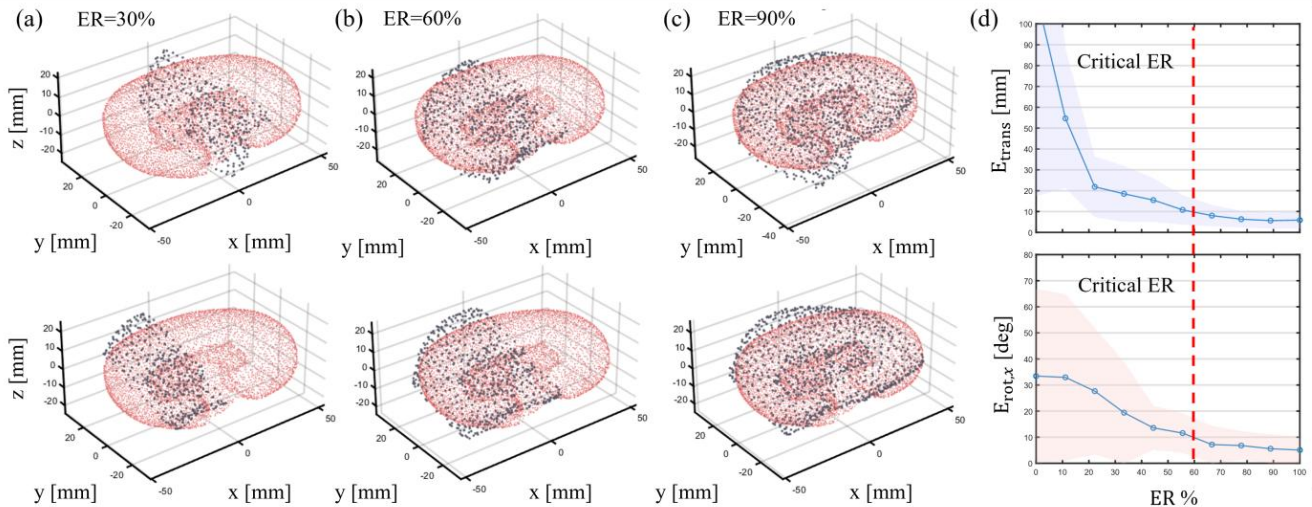

**Fig. 7.** Simulation evaluation on exploration coverage vs. kidney localization accuracy. **(a-c)** Example cases of registering partial exploration kidney point cloud (gray points) to the template kidney point cloud (red points), with ER=30%, 60%, 90%, respectively. Top: estimated registration; Bottom: ground truth registration. **(d)** Kidney localization accuracy vs. exploration ratio. Top: translational accuracy; Bottom: rotational accuracy. The shaded area reveals the standard deviation across patients.

Upon completion of the exploration phase, we compute the estimated kidney pose $T_{k,\text{est}}^b$, then compare it with the ground truth pose $T_{k,\text{gt}}^b$. Same as the previous experiment, $E_{\text{trans}}$ and $E_{\text{rot,x}}$ is employed to quantify the absolute localization error. Moreover, we introduce the US image quality metric: Let the robot move the probe to the optimal pivoting poses $T_{\text{gt}}^*$ and $T_{\text{est}}^*$ using $T_{k,\text{gt}}^b$ and $T_{k,\text{est}}^b$, respectively.

### E. Validate Optimal Pivoting Efficiency in Simulation

The goal of this experiment is to demonstrate the advantage of the proposed optimal pivoting strategy in terms of scanning efficiency. Specifically, we aim to show that our method completes full organ coverage with superior kidney coverage gain per unit probe motion. To this end, the following kidney sweeping methods are compared:

- Proposed optimal pivoting (**OP**): The pivoting is executed at the optimal pivoting pose $T_{\text{est}}^*$.
- Ground truth optimal pivoting (**GT-OP**): The pivoting is executed at the ground truth optimal pivoting pose $T_{\text{gt}}^*$.
- Straight-line sweep (**SL**): A translational sweep is performed to cover the entire kidney. This is equivalent to setting ER = 100% at exploration.
- Non-optimal pivoting (**NOP**): A pivoting is executed to sweep over the transverse planes of the kidney. This is done by aligning the transducer array to the $y$-axis of the kidney canonical frame. The probe tip is located at the projected kidney centroid on the body.

For each virtual patient, we measure the efficiency using two metrics that reflect the cost of whole organ coverage. Namely, the equivalent imaging window path length ($\varepsilon$), and the total US frame count ($n$). $\varepsilon$ is employed to measure the total displacement of the US imaging window during the sweep, computed as:

$$\varepsilon = \sum \sqrt{\|p_t\|^2 + (L\theta_t)^2} \qquad (24)$$

where $p_t$ is the probe's translational displacement; $\theta_t$ is the probe's geodesic rotational displacement; $L$ is the depth of the US image. $n$ is the total number of US frames captured during the sweep. We further evaluate these two metrics against the volume reconstruction ratio $V$, defined as:

$$V = \frac{V_i}{V_{\text{total}}} \qquad (25)$$

where $V_i$ is the kidney volume measured at the $i$-th simulation step. $V_{\text{total}}$ is the total kidney volume. $\varepsilon$, $n$, and $V$ are recorded when the probe is pivoting from one pole of the kidney to the other.

### F. Validate Kidney Localization Accuracy In-vivo

The objective of this experiment is to validate if the proposed framework is robust enough to real-world disturbances (e.g., with tissue deformation, noisy US data, etc.) and still able to accurately localize the kidney.

For each subject, the probe is manually placed on the abdomen where the kidney transverse plane is visible. The exploration phase follows the protocol described in section II-E. In the meantime, paired probe pose and kidney segmentation mask (obtained using YOLO11n-seg model) are recorded. Because the subject's kidney shape is unknown, we cannot calculate the precise exploration path length that corresponds to the critical ER. Instead, we set the same path length for all subjects to be ER $\times L_m$, where $L_m = 100$ mm is the average human kidney length [40]. The kidney localization phase is executed based on explorative imaging. The probe is then moved to the estimated optimal pivoting pose $T^*_{\text{est}}$. Next, the ground truth optimal pivoting pose is determined by letting a human sonographer manipulate the probe to image the kidney's long-axis view. This manual probe placement is repeated three times, and the average probe pose is used as $T^*_{\text{gt}}$. The localization accuracy is evaluated using the same metrics in section III-D, i.e., $E_{\text{trans}}$ and $E_{\text{rot,x}}$.

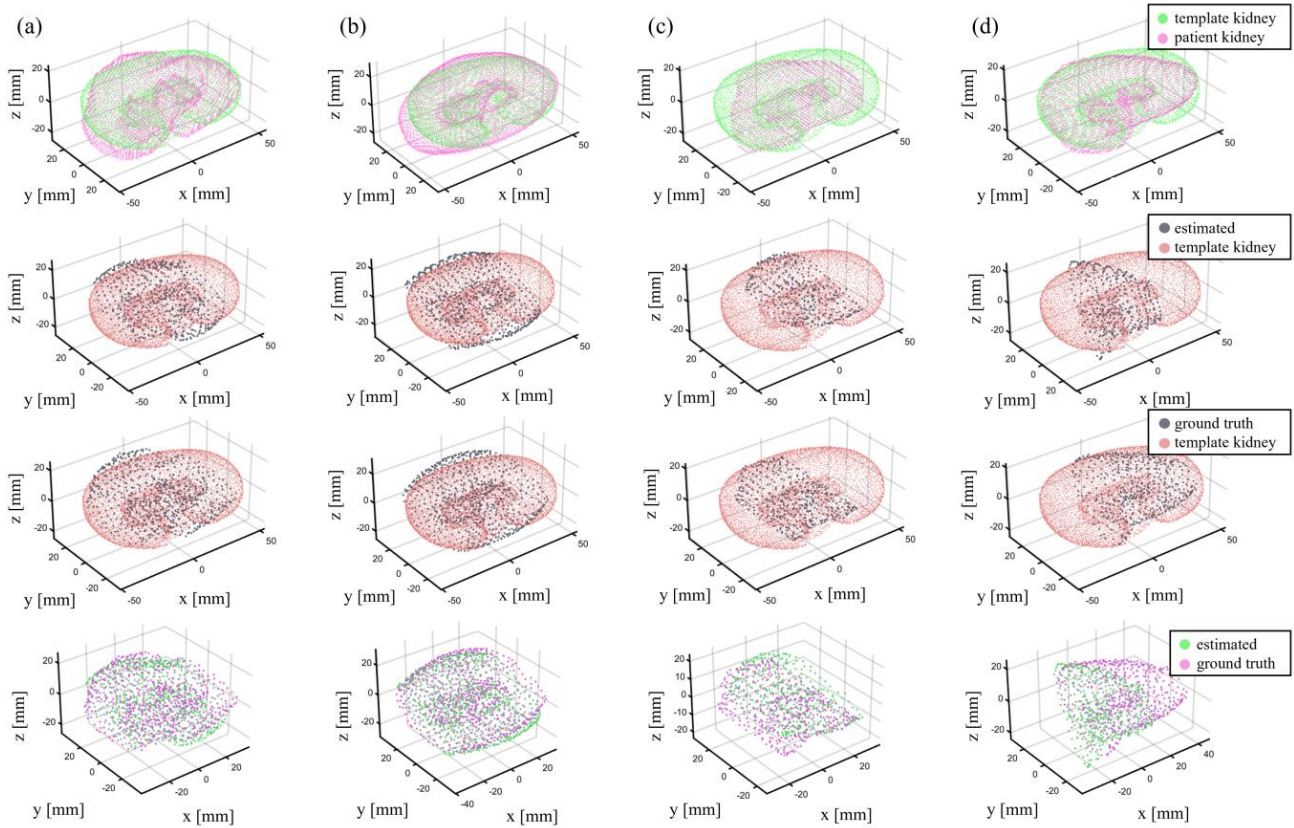

**Fig. 8.** Simulation evaluation of kidney registration accuracy (ER=60%). **(a-b)** Representative "good" cases of kidney registration. Top: estimated registration; Bottom: ground truth registration. **(c-d)** Representative "bad" cases of kidney registration. First row: patient's true kidney shape and the template kidney plotted in the canonical coordinate frame; Second row: estimated kidney registration. Third row: ground truth kidney registration. Fourth row: difference between the estimated and the ground truth local kidney poses with respect to the canonical frame.

### G. Validate Optimal Pivoting Efficiency In-vivo

The objective of this experiment is to demonstrate that the efficiency advantage of the proposed optimal pivoting strategy exists not only in simulation but in realistic clinical settings.

To this end on, we compare the **OP**, **GT-OP**, and the **NOP** strategies aforementioned in section III-E on human subjects. As the superior portion of the kidney is typically hidden beneath the rib cage, it is challenging to form a linear sweep that covers the entire kidney. For this reason, the **SL** sweep is excluded from the comparison.

For each subject, the **OP** sweep is performed using the estimated optimal pivoting pose from section III-F. The **GT-OP** and the **NOP** sweeps are executed by letting a human sonographer place the probe to image the kidney's long-axis view and short-axis view, respectively. We adopted the same efficiency metrics, namely, equivalent imaging window path length ($\varepsilon$) and the total US frame count ($n$). However, because the ground truth kidney shape is unavailable, the total kidney volume is unknown. Therefore, we use the point cloud generated from the **GT-OP** sweep as the proxy to $V_{\text{total}}$ to calculate the volume reconstruction ratio, $V$.

## IV. RESULTS

This section presents the results of the experiments conducted in the previous section, including the exploration coverage vs. kidney localization accuracy evaluation, the kidney localization accuracy evaluation using the critical ER in simulation and in-vivo, the optimal pivoting efficiency evaluation in simulation and in-vivo.

### A. Evaluation of Exploration Coverage vs. Kidney Localization Accuracy

Fig. 7 illustrates the relationship between the Exploration

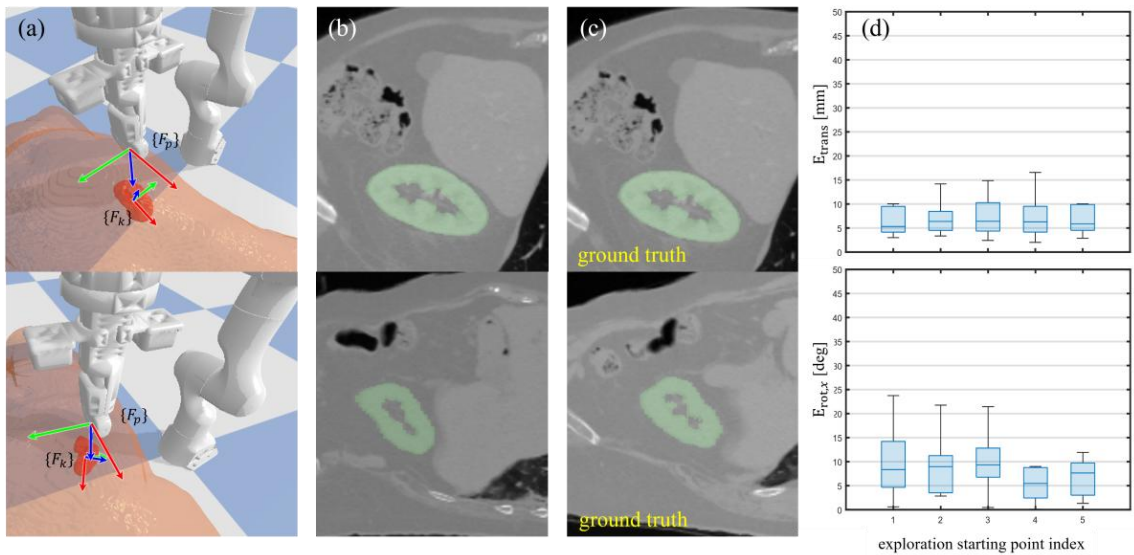

**Fig. 9.** Simulation evaluation of kidney localization accuracy. **(a)** Representative cases of optimal pivoting pose estimations. Top: a "good" case example using kidney registration result from **Fig. 8b** (rotation error: 2.18 degrees); Bottom: a "bad" case example using kidney registration result from **Fig. 8c** (rotation error: 11.33 degrees). **(b)** Emulated US images acquired from the estimated optimal pivoting poses. Top: the good case example; Bottom: the bad case example. **(c)** Emulated US images from the ground truth optimal pivoting poses. Top: the good case example; Bottom: the bad case example. **(d)** Box plot of the kidney localization accuracy in translation (top) and rotation (bottom) at different exploration starting points. The range shows from inter-patient variability.

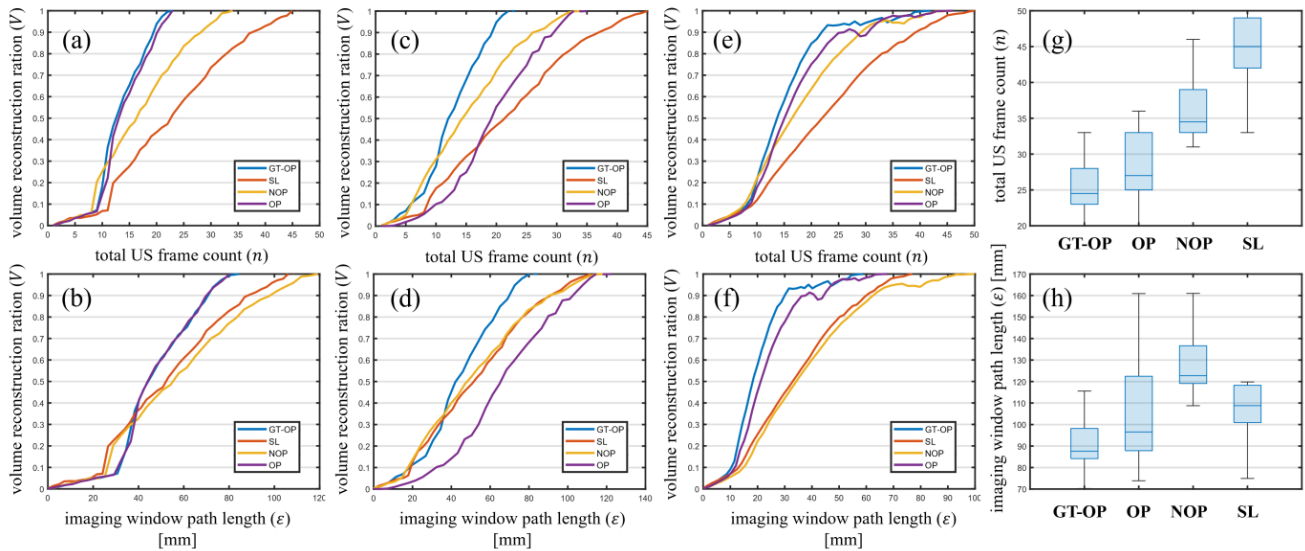

**Fig. 10.** Simulation evaluation of imaging efficiency using different kidney sweeping strategies. **(a-f)** Imaging efficiency evaluated in volume reconstruction ratio vs. total US frame count (top) and imaging window path length (bottom), respectively. **(a-b)** A "good" case example using kidney registration results from **Fig. 8b** (rotation error: 2.18 degrees). **(c-d)** A "bad" case example using kidney registration results from **Fig. 8c** (rotation error: 11.33 degrees). **(e-f)** Average imaging efficiency evaluation across all 15 subjects. **(g-h)** Box plot showing the total US frame count and imaging window path length for the four sweeping strategies.

Ratio (ER) and the precision of the estimated kidney pose. As shown in Fig. 7a-c, a lower ER value (e.g., 30%) yields sparse kidney point cloud that lacks sufficient geometric constraints, leading to visible misalignment when registered to the template kidney. As ER increases, the local point cloud captures enough of the kidney surface curvature and allows more accurate registration to the global template. In the meantime, the variance of the errors across patients also reduces significantly.

Fig. 7d reveals a distinct phase transition in the kidney localization stability with respect to different ER levels. To determine the critical ER, a paired sample t-test is performed, comparing the translation and rotation errors ($E_{trans}$ and $E_{rot,x}$) at each ER to the errors at ER = 100%. The result shows that at ER = 60%, both errors are not significantly different (i.e., p > 0.05) from 100% exploration (p = 0.0744 for $E_{trans}$; p = 0.3523 for $E_{rot,x}$). Hence, we determined that 60% exploration ratio is needed to achieve stable kidney localization. At the critical ER, the average translation error is 10.81 ± 7.08 mm, and the average rotation error is 11.58 ± 7.61 degrees. We used

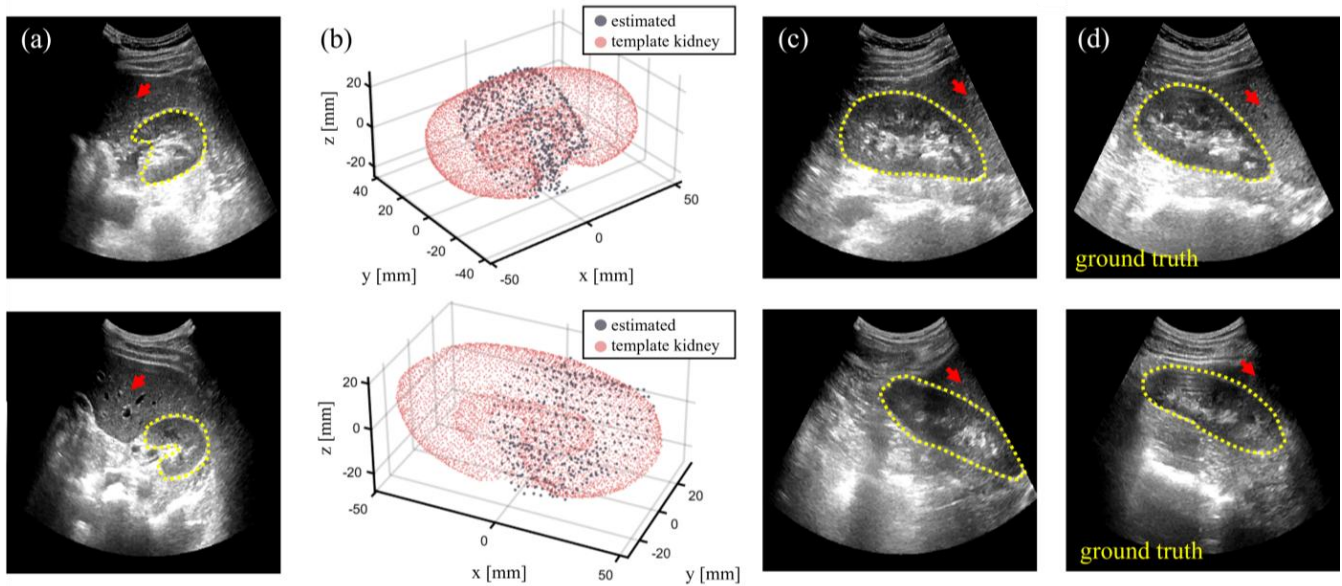

**Fig. 11.** In-vivo evaluation of kidney localization accuracy. **(a)** Example US images acquired during the exploration phase. **(b)** Local kidney point cloud registered to the template kidney. **(c)** US images acquired at the estimated optimal pivoting pose. **(d)** US images acquired at the ground truth optimal pivoting pose. Top: subject 1; Bottom: subject 2. The yellow dotted line highlights the segmented kidney contour. The red arrows show the liver which is an important anatomical landmark for right kidney imaging.

ER=60% for all subsequent experiments.

The above findings align closely with our theoretical expectation that the partial-to-global registration problem is highly underdetermined. When only a small fraction of the kidney is observed, the local point cloud is geometrically symmetrical enough to allow the ICP algorithm to slide or tilt within the local minima. Even though the kidney exhibits an overall smooth surface, the renal poles and the central concave curvature are the primary drivers of the localization. These features typically only become prominent once the probe has transversed more than half of the organ.

### B. Evaluation of Kidney Localization Accuracy in Simulation

Here, we show the kidney localization accuracy evaluation using the finalized 60% ER. Fig. 8 and Fig. 9 collectively demonstrate the system's ability to solve for the organ pose from local kidney exploration.

Fig. 8a-b shows two good cases of a nearly seamless registration of the partial kidney to the template. The translation errors ($E_{trans}$) for these two cases are 3.84 mm and 3.12 mm, respectively. The rotation errors ($E_{rot,x}$) are 1.24 degrees and 2.18 degrees, respectively. Fig. 8c-d shows two bad cases with relatively greater registration errors. The translation errors for these two cases are 11.33 mm and 13.41 mm, respectively. The rotation errors are 10.80 degrees and 13.81 degrees, respectively. The bad cases are largely caused by the significant anatomical deformation from the kidney template model (e.g., when the kidney is noticeably smaller in volume). Among all 15 subjects, the average translation error is $8.73 \pm 6.93$ mm, while the average rotation error is $12.86 \pm 20.86$ degrees. These statistics suggest that the main bottleneck for the kidney localization accuracy is the inter-patient anatomical

discrepancy. Nonetheless, the overall translational error is relatively small, allowing the kidney to appear within the US imaging FOV (around 10 cm lateral width at 10 cm depth). On the other hand, the higher rotational error may result in longer pivoting period.

Fig. 9d shows the kidney localization error at different starting points during exploration. To investigate whether the kidney localization accuracy is affected by the specific area being explored, we performed a paired t-test among the five randomized exploration starting points (10 t-tests in total). Using p > 0.005 (Bonferroni correction) as the significance threshold, we found no significant differences in both translation and rotation errors. This indicates that the exploration starting point does not affected the final localization accuracy – the 60% critical exploration ratio can produce consistent kidney pose estimation.

Fig. 9a-c further visualizes the effect of kidney localization error on the optimal pivoting pose. A "good" localization results in a long-axis view nearly identical to the ground truth, where the kidney is at the center of the image and the length of the organ is more accurately revealed.

In summary, the simulation study successfully demonstrated a sub-centimeter kidney localization accuracy. While rotation variance exists, the system can still identify the longitudinal axis well enough to provide a viable imaging window for subsequent volumetric sweep.

### C. Evaluation of Optimal Pivoting Efficiency in Simulation

The simulation evaluation of the kidney imaging efficiency is shown in Fig. 10.

Fig. 10a-b demonstrates the volume reconstruction ratio $V$ curve with respect to the US frame count ($n$) and the imaging

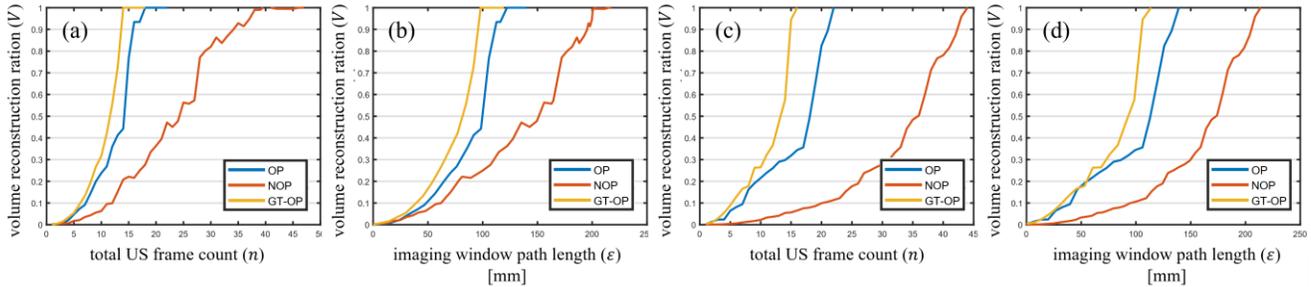

**Fig. 12.** In-vivo evaluation of imaging efficiency using different kidney sweeping strategies. **(a)** Volume reconstruction ratio vs. total US frame count for subject 1. **(b)** Volume reconstruction ratio vs. imaging window path length for subject 1. **(c)** Volume reconstruction ratio vs. total US frame count for subject 2. **(d)** Volume reconstruction ratio vs. imaging window path length for subject 2.

window path length ($\varepsilon$) when the kidney localization is relatively accurate ($E_{trans}$ =3.12 mm, $E_{rot,x}$ =2.18 degrees). Under such conditions, the proposed optimal pivoting strategy (**OP**) exhibits steeper curve compared to the non-optimal pivoting (**OP**) and the straight-line sweep (**SL**). This indicates that the **OP** strategy achieves full kidney coverage with the probe traveled less, and the kidney volume gain per US image is higher, hence the most efficient.

Fig. 10c-d shows the same $V$ curves when the kidney localization is less accurate ($E_{trans}$ =11.33 mm, $E_{rot,x}$ =10.80 degrees). In this case, the **OP** strategy leads to slower kidney volume increase because the imaging plane is deviated from the kidney's longitudinal plane. The difference of the non-optimal pivoting and the **GT-OP** strategy also reflects this deviation. Despite the larger kidney localization error, the **OP** strategy still outperforms the **SL** strategy in terms of US frame count. The imaging window travels an approximately equal amount of distance to the SL and NOP strategies.

Fig. 10e-f shows the average $V$ curves across all 15 subjects, which demonstrate the overall advantages of the proposed **OP** strategy in terms of imaging efficiency. Fig. 10g-h further quantifies the advantage over the **SL** strategy and the **NOP** strategy. On average, the **OP** strategy reduces the imaging plane displacement by $1.71 \pm 29.62$ mm compared to the **SL** strategy and by $24.66 \pm 21.50$ mm compared to the **NOP** strategy. Meanwhile, the imperfect kidney localization results in an extended imaging plane displacement of $11.71 \pm 12.60$ mm.

In summary, the proposed **OP** strategy achieves whole kidney coverage using less US frames and imaging window displacement then the baseline strategies. The result aligns with our intuition that long-axis pivoting is the most efficient geometric approach for kidney scanning. The reduction in the imaging window displacement is a valuable advantage as the available acoustic window for kidney imaging is often limited in clinical practice.

### D. Evaluation of Kidney Localization In-vivo

The in-vivo result of the kidney localization accuracy evaluation is illustrated in Fig. 11. A full-workflow demonstration can also be found in the **supplementary video**. Fig. 11a shows that high quality US images with clear anatomies can be captured during the exploration phase. Fig.

11b demonstrates the local kidney point cloud registered to the template model. There is no major jittering observed in the local point cloud, underscoring the robustness of the kidney segmentation model. The point cloud alignment further confirms that a 60% ER scan provides sufficient geometric information to lock the local kidney shape to the global template.

Because the ground truth kidney pose is un available for in-vivo experiments, the kidney localization accuracy is measured as the position ($E_{trans}$) and orientation ($E_{rot,x}$) discrepancy at the estimated optimal pivoting pose relative to the manual expert probe placement. For subject 1, $E_{trans}$ =7.36 mm, $E_{rot,x}$ =13.84 degrees. For subject 2, $E_{trans}$ =16.96 mm, $E_{rot,x}$ =30.57 degrees. Despite the higher numerical values than the simulation study, the resulting US images (see Fig. 11c) still slice the longitudinal sections of the kidney, mirroring the expert's targeted view (see Fig. 11d).

As expected, the kidney localization error appears to be larger in the in-vivo study compared to the simulation. This gap is most likely due to the following reasons: (i) Slight body movement may still exist although we attempted to avoid it in the experiment protocol; (ii) The probe is placed firmly into the body, causing noticeable tissue deformation that changes the kidney pose. Nevertheless, the results demonstrate the overall effectiveness of our proposed "exploration—localization" workflow. Moreover, the captured kidney length at the estimated and ground truth optimal pivoting pose is approximately the same. By identifying a probe pose that captures the majority of the longitudinal axis, rather than the short axis views during exploration, the system demonstrates strong potential for autonomous diagnosis assistance.

### E. Evaluation of Optimal Pivoting Efficiency In-vivo

Fig. 12 presents the imaging efficiency for the two human subjects across three kidney sweeping strategies: **OP**, **GT-OP** (expert-guided long axis pivoting), and **NOP** (expert-guided short axis pivoting). As mentioned in section III-G, the **SL** strategy is excluded because a straight-line trajectory covering whole kidney is not available.

For both subjects, the **OP** and **GT-OP** strategy achieve full kidney imaging significantly faster than the NOP method. The wider gap between **OP** and **GT-OP** strategy for subject 2 is explained by the larger kidney localization error. Using the proposed **OP** strategy, the total imaging plane path length is reduced by 79 mm and 73 mm, respectively. Notice that for subject 2, the advantage of the **OP** over **NOP** is more significant. This is likely because subject 2 possesses a relatively thin and elongated kidney (see Fig. 11d), hence the probe needs to travel wider angular range during short-axis pivoting.

The in-vivo result shows similar trend of the $V$ curve, reaffirming that the optimal pivoting strategy is more efficient than the baseline methods.

## V. CONCLUSION AND DISCUSSION

This paper presents a novel autonomous workflow for kidney US to achieve more efficient organ coverage than the traditional freehand and robotic scanning methods. The proposed approach features a template-guided exploration-localization framework, involving three progressive phases: the initial explorative imaging phase, the kidney localization phase, and the optimal pivoting phase. Through this approach, RUS systems can identify the geometrically optimal imaging window for volumetric sweep without the need of prior, patient-specific information. The significance of the proposed framework lies in two aspects: (i) it achieves global organ pose estimation from partial observations, which is fundamental to enable autonomous, standardized imaging for RUS systems; (ii) it demonstrates spatially efficient imaging by minimizing the probe footprint while maximizing the organ coverage, which is critical to imaging organs (e.g., kidney) located deep inside the body with limited acoustic window. Our results show that the proposed method effectively localizes the kidney with approximately 10 mm accuracy and reduces the imaging window path length (i.e., probe footprint) by 73-79 mm on human subjects using the optimal pivoting strategy.

### A. Implications from the Results

The simulation and in-vivo experiment results provide several key takeaways regarding the robustness and the efficiency of the system. In the simulation study involving 15 virtual patients, we found that (i) a 60% exploration ratio (ER) is the critical threshold that balances the localization accuracy and the procedural efficiency, and (ii) with 60% ER, the difference in the explorative imaging area does not significantly affect the kidney localization accuracy. Furthermore, we demonstrated that a kidney localization accuracy of $8.73 \pm 6.93$ mm and $12.86 \pm 20.86$ degrees can be achieved in simulation. This error is primarily attributed to inter-subject anatomical variability. Yet the accuracy is sufficient to ensure the visualization of the kidney's long axis plane at the optimal pivoting pose. The subsequent efficiency test demonstrates that the proposed optimal pivoting (**OP**) strategy consistently outperforms the straight-line sweep (**SL**) and non-optimal sweep (**NOP**) baselines. While the simulation shows high kidney localization precision, the in-vivo evaluation exhibited larger localization errors (7.36 to 16.96 mm and 13.84 to 30.57 degrees). This performance gap is mostly likely due to slight kidney pose changes caused by involuntary movement and tissue deformation, as well as the use of expert manual identification as the ground truth. Nonetheless, the system is still able to capture the kidney inside the image FOV and demonstrate superior imaging efficiency using our **OP** strategy, suggesting the overall effectiveness of the proposed framework. Notice that while a sub-millimeter accuracy is desired, there are methods that can potentially further reduce the error. For instance, one could involve US image feedback to center the kidney in the image and perform local probe maneuver to optimize the long-axis view. However, such image-based fine-tuning methods are beyond the scope of this work and have been explored by other groups (e.g., [11]). In this study, we are focusing more on demonstrating the feasibility of the exploration-localization workflow for autonomous imaging.

### B. Generalizability of the Proposed Framework

Beyond the specific application of renal US, our exploration-localization concept can be used as a general-purpose paradigm in the field of robot-assisted medical imaging. At its core, our method addresses the partial-to-global registration problem, which is fundamental to many robotic imaging procedures where the sensor FOV is limited (e.g., US, OCT [41], and photoacoustic tomography [42]). By contributing a PCA-based canonical template approach, this paper provides a roadmap for how robots can "infer" the global pose of an organ or anatomy from minimum initial data. We choose kidney as the primary organ model because it represents a "worst case" benchmark: it is located with narrow acoustic window from the outside of the body; its surface is smooth without significant landmarks. Success in kidney imaging implies that the system is robust enough to handle simpler, landmark-rich anatomies. By replacing the kidney PCA model with other anatomical templates, the proposed workflow could facilitate standardized, autonomous volumetric scans for other imaging tasks such as spine [43] and heart [44].

### C. Limitations and Future Directions

Despite the promising results, several limitations remain. First, the current workflow assumes a relatively static environment. However, patient body movement and soft tissue deformation are often unavoidable in practice. Second, the template kidney is currently a rigid body model, which does not account for the morphological variations in patients with renal diseases. These two factors may result in degraded kidney localization performance. Finally, the in-vivo validation was conducted on a limited cohort, which does not fully capture the spectrum of anatomical anomalies in the population.

To address these limitations, future work will focus on integrating real-time tracking of the patient pose and the respiration phases. Deformable template model will be investigated to restore the true shape of the patient's kidney from partial observations for better registration accuracy. Additionally, conducting a larger scale clinical study and expanding the system to support multi-organ imaging protocols will be essential steps towards deploying this technology in diagnostic environments. Ultimately, this research paves the way for fully autonomous, anatomy-aware RUS systems capable of delivering high quality volumetric imaging with minimum human intervention.

REFERENCES

[1] P. R. Hoskins, K. Martin, and A. Thrush, Eds., Diagnostic Ultrasound. Third edition. | Boca Raton, FL: CRC Press/Taylor & Francis Group, [2019]: CRC Press, 2019.

[2] K. J. Jager, C. Kovesdy, R. Langham, M. Rosenberg, V. Jha, and C. Zoccali, "A single number for advocacy and communication—worldwide more than 850 million individuals with kidney diseases," Kidney International, vol. 96, no. 5, pp. 1048–1050, Nov. 2019.

[3] R. K. Singla, M. Kadatz, R. Rohling, and C. Nguan, "Kidney Ultrasound for Nephrologists: A Review," Kidney Medicine, vol. 4, no. 6, p. 100464, Jun. 2022.

[4] M. Mahboob, P. Rout, and Syed, "Autosomal Dominant Polycystic Kidney Disease," Nih.gov, Mar. 20, 2024.

[5] Q. Zhou et al., "Current Status of Ultrasound in Acute Rejection After Renal Transplantation: A Review with a Focus on Contrast-Enhanced Ultrasound," Annals of Transplantation, vol. 26, Apr. 2021.

[6] S. J. Frank, W. R. Walter, L. Latson, H. W. Cohen, and M. Koenigsberg, "New Dimensions in Renal Transplant Sonography," Transplantation, vol. 101, no. 6, pp. 1344–1352, Jun. 2017.

[7] J. M. Jagtap et al., "Automated measurement of total kidney volume from 3D ultrasound images of patients affected by polycystic kidney disease and comparison to MR measurements," Abdominal Radiology, vol. 47, no. 7, pp. 2408–2419, Apr. 2022.

[8] J. Bakker et al., "Renal Volume Measurements: Accuracy and Repeatability of US Compared with That of MR Imaging," vol. 211, no. 3, pp. 623–628, Jun. 1999.

[9] M. Dhyani et al., "A pilot study to precisely quantify forces applied by sonographers while scanning: A step toward reducing ergonomic injury," Work, vol. 58, no. 2, pp. 241–247, Oct. 2017.

[10] Z. Jiang, S. E. Salcudean, and Nassir Navab, "Robotic ultrasound imaging: State-of-the-art and future perspectives," Medical Image Analysis, vol. 89, pp. 102878–102878, Jul. 2023.

[11] Z. Jiang et al., "Automatic Normalization of Robotic Ultrasound Probe Based Only on Confidence Map Optimization and Force Measurement," IEEE Robotics and Automation Letters, vol. 5, no. 2, pp. 1342–1349, Jan. 2020.

[12] Q. Huang, J. Lan, and X. Li, "Robotic Arm Based Automatic Ultrasound Scanning for Three-Dimensional Imaging," IEEE Transactions on Industrial Informatics, vol. 15, no. 2, pp. 1173–1182, Sep. 2018.

[13] R. Tsumura et al., "Tele-Operative Low-Cost Robotic Lung Ultrasound Scanning Platform for Triage of COVID-19 Patients," IEEE Robotics and Automation Letters, vol. 6, no. 3, pp. 4664–4671, Jul. 2021.

[14] L. Lindenroth, R. J. Housden, S. Wang, J. Back, K. Rhode, and H. Liu, "Design and Integration of a Parallel, Soft Robotic End-Effector for Extracorporeal Ultrasound," IEEE Transactions on Biomedical Engineering, vol. 67, no. 8, pp. 2215–2229, Aug. 2020.

[15] X. Ma, W.-Y. Kuo, K. Yang, A. Rahaman, and H. K. Zhang, "A-SEE: Active-Sensing End-Effector Enabled Probe Self-Normal-Positioning for Robotic Ultrasound Imaging Applications," IEEE Robotics and Automation Letters, vol. 7, no. 4, pp. 12475–12482, Oct. 2022.

[16] Y. Zhetpissov, X. Ma, K. Yang, and H. K. Zhang, "A-SEE2.0: Active-Sensing End-Effector for Robotic Ultrasound Systems with Dense Contact Surface Perception Enabled Probe Orientation Adjustment," IEEE Robotics and Automation Letters, vol. 10, no. 9, pp. 9557–9564, Sep. 2025.

[17] P. Abolmaesumi, S.E. Salcudean, W.-H. Zhu, M.R. Sirouspour, and S. P. DiMaio, "Image-guided control of a robot for medical ultrasound," IEEE Transactions on Robotics and Automation, vol. 18, no. 1, pp. 11–23, Jan. 2002.

[18] C. Nadeau and A. Krupa, "Intensity-based direct visual servoing of an ultrasound probe," 2011 IEEE International Conference on Robotics and Automation, Shanghai, China, 2011, pp. 5677-5682.

[19] R. Mebarki, A. Krupa, and F. Chaumette, "2-D Ultrasound Probe Complete Guidance by Visual Servoing Using Image Moments," IEEE Transactions on Robotics, vol. 26, no. 2, pp. 296–306, Apr. 2010.

[20] X. Ma, M. Zeng, J. C. Hill, B. Hoffmann, Z. Zhang, and H. K. Zhang, "Guiding the Last Centimeter: Novel Anatomy-Aware Probe Servoing for Standardized Imaging Plane Navigation in Robotic Lung Ultrasound," IEEE Transactions on Automation Science and Engineering, vol. 22, pp. 6569–6580, 2025.

[21] K. Li, Y. Xu, J. Wang, D. Ni, L. Liu, and M. Q.-H. . Meng, "Image-Guided Navigation of a Robotic Ultrasound Probe for Autonomous Spinal Sonography Using a Shadow-Aware Dual-Agent Framework," IEEE Transactions on Medical Robotics and Bionics, vol. 4, no. 1, pp. 130–144, Feb. 2022.

[22] G. Ning, X. Zhang, and H. Liao, "Autonomic Robotic Ultrasound Imaging System Based on Reinforcement Learning," IEEE Transactions on Biomedical Engineering, vol. 68, no. 9, pp. 2787–2797, Jan. 2021.

[23] Y. Bi, C. Qian, Z. Zhang, N. Navab, and Z. Jiang, "Autonomous Path Planning for Intercostal Robotic Ultrasound Imaging Using Reinforcement Learning," arXiv.org, 2024.

[24] Z. Jiang, Y. Bi, M. Zhou, Y. Hu, M. Burke, and Nassir Navab, "Intelligent robotic sonographer: Mutual information-based disentangled reward learning from few demonstrations," The International Journal of Robotics Research, vol. 43, no. 7, pp. 981–1002, Jan. 2024.

[25] Z. Jiang et al., "Precise Repositioning of Robotic Ultrasound: Improving Registration-Based Motion Compensation Using Ultrasound Confidence Optimization," IEEE Transactions on Instrumentation and Measurement, vol. 71, pp. 1–11, Jan. 2022.

[26] J. Zielke, C. Eilers, B. Busam, W. Weber, N. Navab, and T. Wendler, "RSV: Robotic Sonography for Thyroid Volumetry," IEEE Robotics and Automation Letters, vol. 7, no. 2, pp. 3342–3348, Apr. 2022.

[27] F. Suligoj, C. M. Heunis, J. Sikorski, and S. Misra, "RobUSt–An Autonomous Robotic Ultrasound System for Medical Imaging," IEEE Access, vol. 9, pp. 67456–67465, 2021.

[28] M. Victorova, H. H. T. Lau, T. T. Lee, D. Navarro-Alarcon, and Y. Zheng, "Comparison of ultrasound scanning for scoliosis assessment: Robotic versus manual," The International Journal of Medical Robotics and Computer Assisted Surgery, vol. 19, no. 2, Nov. 2022.

[29] X. Ma, Z. Zhang, and H. K. Zhang, "Autonomous Scanning Target Localization for Robotic Lung Ultrasound Imaging," 2021 IEEE/RSJ International Conference on Intelligent Robots and Systems (IROS), pp. 9467–9474, Sep. 2021.

[30] K. Su et al., "A fully autonomous robotic ultrasound system for thyroid scanning," Nature Communications, vol. 15, no. 1, pp. 4004–4004, May 2024.

[31] L. Lei et al., "Toward Lung Ultrasound Automation: Fully Autonomous Robotic Longitudinal and Transverse Scans Along Intercostal Spaces," IEEE Transactions on Medical Robotics and Bionics, vol. 7, no. 2, pp. 768–781, May 2025.

[32] J. Elsner, "Taming the Panda with Python: A powerful duo for seamless robotics programming and integration," SoftwareX, vol. 24, p. 101532, Dec. 2023.

[33] Jakob Wasserthal et al., "TotalSegmentator: Robust Segmentation of 104 Anatomic Structures in CT Images," Radiology Artificial Intelligence, vol. 5, no. 5, Jul. 2023.

[34] X. Ma et al., "Cross-Modality Registration using Bone Surface Pointcloud for Robotic Ultrasound-Guided Spine Surgery," Journal of Medical Robotics Research, vol. 10, no. 01n02, Dec. 2024.

[35] G. Jocher and J. Qiu, Ultralytics YOLO11. 2024. [Online]. Available: https://github.com/ultralytics/ultralytics.

[36] K. kaur, "Kidney Ultrasound Images 'Stone' and 'No Stone,'" Kaggle.com, 2024. https://www.kaggle.com/datasets/gurjeetkaurmangat/kidney-ultrasound-images-stone-and-no-stone.

[37] J. Ma, Y. He, F. Li, L. Han, C. You, and B. Wang, "Segment anything in medical images," Nature Communications, vol. 15, no. 1, Jan. 2024.

[38] X. Ma, Yemar Zhetpissov, and H. K. Zhang, "RUS-Sim: A Robotic Ultrasound Simulator Modeling Patient-Robot Interaction and Real-Time Image Acquisition," pp. 1–4, Sep. 2024.

[39] K. Hansen, M. Nielsen, and C. Ewertsen, "Ultrasonography of the Kidney: A Pictorial Review," Diagnostics, vol. 6, no. 1, pp. 2–2, Dec. 2015.

[40] Wael El-Reshaid and Husam Abdul-Fattah, "Sonographic Assessment of Renal Size in Healthy Adults," Medical Principles and Practice, vol. 23, no. 5, pp. 432–436, Jan. 2014.

[41] X. Ma et al., "Large area kidney imaging for pre-transplant evaluation using real-time robotic optical coherence tomography," Communications Engineering, vol. 3, no. 1, May 2024.

[42] S. Gao, X. Ma and H. K. Zhang, "Robot-Assisted Wide-Area Photoacoustic System," 2023 IEEE International Ultrasonics Symposium (IUS), Montreal, QC, Canada, 2023, pp. 1-4.

[43] L. Tang, Z. Hu, Y.-S. Lin, and J. Hu, "A statistical lumbar spine geometry model accounting for variations by Age, Sex, Stature, and body mass index," Journal of Biomechanics, vol. 130, pp. 110821–110821, Oct. 2021.

[44] C. Rodero et al., "Linking statistical shape models and simulated function in the healthy adult human heart," PLOS Computational Biology, vol. 17, no. 4, p. e1008851, Apr. 2021.